# GLOBAL DISASTER RISK ASSESSMENT USING WORLD RISK INDEX DATA: A GEOSPATIAL AND VISUAL ANALYTICS APPROACH

Malithi Nawarathne
*Faculty of Computing, Sri Lanka Institute of Information Technology, Sri Lanka*

Nadeesha Herath Mudiyanselage
*Department of Computer Science, Faculty of Science, University of Helsinki, Finland*

Lihini Sangeetha
*Computing Centre, Faculty of Engineering, University of Peradeniya, Sri Lanka*

---

## ABSTRACT

*The world has seen several threats that include resource depletion, food security, poverty and inequality, difficulties associated with urbanization, migration, and environmental effects. The World Risk Index (WRI) is a metric that is used to measure the susceptibility of nations against natural disasters and related risks that is reported by the Humanitarian Data Exchange. The WRI is comprised of factors such as susceptibility, vulnerability to natural hazards, and coping mechanisms. The WRI index has allowed to recognize regions in advance in order to face threats and situations beforehand. This study was conducted by analyzing the data that was related to years from 2000 to 2025 using data visualization techniques. Python related libraries of Matplotlib, Seaborn, and Geopandas were used due to their efficiency and usability in analyzing WRI data. As per the analysis, it was identified that countries of Philippines, Mexico, Indonesia, and India were recorded with high WRI index values that showed these nations should be prepared for natural disasters and situations. Hence, this study has provided insights using data visualization techniques in understanding natural disasters and threats so that necessary measures can be taken.*

**KEYWORDS:** *Data Visualization, Disaster Risk Assessment, Geospatial Analysis, Visual Analytics, World Risk Index*

Corresponding Author: Malithi Nawarathne email: malithi.n@sliit.lk

# 1. INTRODUCTION

The world population has been increasing rapidly over the years (Population Matters, 2024). The population growth has created problems of insufficient food, water, energy and other necessary materials. This issue has created a serious competition for essential resources, food security and caused poverty, urbanization, and creating a strong impact on the environment and climate (Poljansek et al., 2017). . The population growth has also caused the rapid production of food via agricultural products. However, the lack of ample water resources, changes in climate has caused doubts about preserving food security. The rise of population has also posed issues on habitats to build houses that has made the environmental resources to be exhausted. Given these matters, it is important to plan for sustainable living by studying the statistics and factors related to countries.

The World Risk Report 2025 is a report that evaluates the risk of disaster for 193 nations (Humanitarian Data Exchange, 2025). The risk assessment in the World Risk Report is based on the general understanding that the overall state of society is just as important to disaster risk as the intensity of the extreme natural event. The "World Risk Index" (WRI) is a tool used to evaluate and rank each nation's susceptibility to natural disasters and the risks that go along with it. The United Nations University Institute for Environment and Human Security (UNU-EHS) developed the WRI, which is based on several indicators related to societal vulnerability and natural hazard risk. These indicators typically take into account things like population susceptibility, such as poverty rates and access to healthcare, exposure to natural hazards such as earthquakes, floods, and cyclones, and coping mechanisms such as governance and infrastructure, which were also mentioned above. The index is frequently represented by a number, with higher values denoting higher risk levels. The index aids in the prioritization of resources and interventions to lessen vulnerability and boost resilience in regions vulnerable to natural disasters by policymakers, researchers, and organizations.

However, no adequate research has been done in the field of global risk assessment, especially when it comes to the topic of world risk, while Welle and Birkmann (2015) have examined the probability that a natural disaster will occur in 193 distinct nations. The WRI data have been mapped and classified using a Geographic Information System (GIS) method to display different exposure, vulnerability, and risk patterns on two world scales. As the global perspective on risk makes plainly obvious, the vulnerability of a community or a nation is distinct from its exposure to natural hazards, according to the authors' findings. Additionally, the results of this study have shown that risk and vulnerability can be evaluated, and that factors such as a nation's exposure to natural hazards and the effects of climate change, as well as social framework conditions and action-capacity indicators that are reflected in susceptibility, coping capacities, and adaptive capacities, are equally important in determining disaster risk.

A mixed hierarchical fuzzy logic approach for determining the risk index for water supply pipes has been presented in a study by Fayaz et al. (2018). This study has suggested a model based on hierarchical fuzzy logic and a GIS map for evaluating and displaying the water supply risk index. The researchers have demonstrated that the hierarchical fuzzy logic-based invention can be applied in real-world scenarios, where the calculated risk index values of the water supply pipeline are plotted on a map created by a GIS, providing users with a graphical interface.

Furthermore, a group of researchers has presented a model of the variables influencing global human losses from natural disasters between 1980 and 2000 . (Peduzzi et al., 2009). The model assesses the relationship between observed historical losses, population exposure, and vulnerability. The findings of this study showed that environmental quality and country development levels are primarily related to human vulnerability. Moreover, Suhaila (2021) has used the functional principal component technique to find new functions that would reveal the most significant type of variation in the multivariate El Niño Southern Oscillation index curve with the help of graphs such as rainbow graphs, functional bagplots, and functional high-density region boxplots.

Overall, these papers collectively contribute to the field of risk assessment and management. They emphasize the need for comprehensive approaches that consider exposure, vulnerability, and risk factors. Therefore, this study has conducted an analysis using the world risk data to draw insights to understand which countries have been found to be at risk, considering the world risk index. The paper is structured so that section 2 provides the methodology, results and discussion presented in section 3, while section 4 includes the conclusion of the work.

## 2. METHODOLOGY

### 2.1 Dataset

The dataset for this study was retrieved from the World Risk Data Repository, and it contained data for 25 years, ranging from 2000 to 2025, including 193 countries (Humanitarian Data Exchange, 2025). For this study, each region's WRI score, exposure score, vulnerability score, susceptibility score, lack of coping capabilities score, and lack of adaptive capacity score were considered. Figure 1 demonstrates a sample of the original dataset that was used for this study.

**Table 1: World risk dataset description**

| Column name | Description |
|---|---|
| Region | Country name |
| ISO3 | Country code |
| WRI | World risk score associated with the country |
| Vulnerability | Vulnerability considering factors such as nutrition, housing situation, economic conditions, and infrastructure |
| Susceptibility | Susceptibility considering nutrition, housing situation, and economic framework conditions |
| Lack of Coping Capabilities | Coping skills in relation to governance, readiness, and early warning, healthcare, and material and social security. |
| Lack of Adaptive Capabilities | Adaptability to upcoming natural disasters, climate change, and other difficulties. |
| Year | Relevant year |

In addition, a detailed description of each column of the original dataset is depicted in Table 1.

### 2.2 Methodology

The proposed methodology of this study is represented in Figure 2 where the data were analyzed and checked for availability of null values.

**Figure 1: A sample of the original dataset.**

.

| | Region | ISO3 | Year | WorldRiskIndex | Exposure | Vulnerability | Susceptibility | Lack of Coping Capacities | Lack of Adaptive Capacities |
|---|---|---|---|---|---|---|---|---|---|
| **0** | Afghanistan | AFG | 2000 | 4.18 | 0.25 | 69.83 | 61.97 | 73.22 | 75.05 |
| **1** | Albania | ALB | 2000 | 7.23 | 2.38 | 21.96 | 22.58 | 44.37 | 10.57 |
| **2** | Algeria | DZA | 2000 | 10.28 | 2.65 | 39.85 | 29.57 | 52.73 | 40.58 |
| **3** | Andorra | AND | 2000 | 0.29 | 0.02 | 4.32 | 2.84 | 2.15 | 13.18 |
| **4** | Angola | AGO | 2000 | 10.98 | 2.18 | 55.35 | 35.29 | 62.14 | 77.32 |
| **...** | ... | ... | ... | ... | ... | ... | ... | ... | ... |

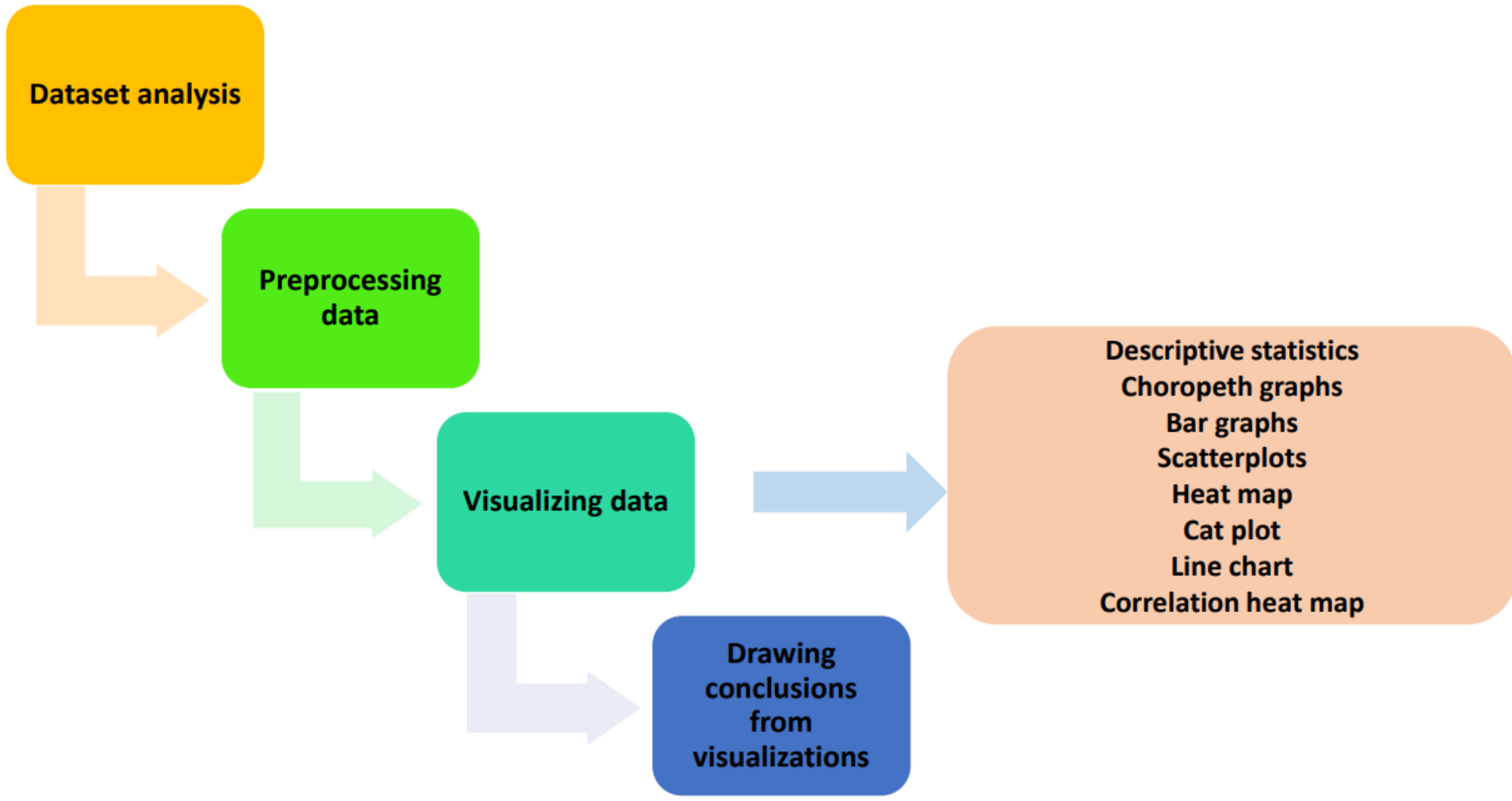

**Figure 2: The methodology of the proposed study.**

It was identified that there were zero rows with missing values. Therefore, without requiring further preprocessing, these data were used for visualizations. The proposed study discusses what factors affect the analysis of the World Risk Index using different data visualization techniques and popular libraries for analyzing WRI data.

### 2.3 Data Visualization

Presenting data in a graphical or pictorial format to make it easier to understand is known as data visualization (Sadiku et al., 2016). It facilitates the decision-making process and the explanation of facts. Therefore, data visualization is a crucial factor in any study. This study used Python-based visualization libraries such as Matplotlib, Seaborn, and Geopandas for visualization. A description of the type of graphs and charts that were used is given below.

**1. Catplot**

Catplots can be made using the Python function catplot, which is part of the Seaborn library. It is a higher-level function that offers a versatile interface for making different kinds of categorical plots. A catplot was used in the proposed study to observe how the WRI, exposure, vulnerability, susceptibility, lack of coping capabilities, and lack of adaptive capabilities have evolved over the recorded years.

**2. Choropleth maps**

The Geopandas library was used to visualize Choropeth maps using the data. Choropeth maps display normalized data as shaded points, lines, or areas, using the counts and amounts (Color) smart mapping symbol type (ArcGIS, n.d.). Choropeth maps provide geographical representations that aid in interpreting visualizations which are explained in Table 2.

**Table 2: Purpose and insights of Choropleth maps**

| **Choropleth maps** | **Purpose** | **Insight** |
|---|---|---|
| WRI by country | To visualize the overall disaster risk level of each country | Identifies which countries are at the highest overall risk of natural disasters.<br><br>Highlights regions that need urgent attention and resource allocation for disaster risk reduction. |
| Exposure by country | To show the level of exposure to natural hazards for each country. | Indicates areas with high likelihood of experiencing natural disasters.<br><br>Helps in understanding geographic and climatic patterns contributing to high exposure.<br><br>Guides infrastructure planning and development to mitigate hazard exposure. |
| Vulnerability by country | To depict the vulnerability levels of countries to natural hazards. | Reveals how susceptible countries are to the impacts of disasters, irrespective of their exposure levels.<br><br>Highlights countries that need to improve their health systems, economic stability, and social structures to reduce vulnerability. |
| Susceptibility by country | To illustrate the susceptibility of countries to damage and harm from natural hazards. | Identifies countries where physical, economic, and social conditions make them prone to severe damage from disasters.<br><br>Helps in targeting efforts to strengthen infrastructure and social safety nets. |
| Lack of Coping Capabilities by Country | To map the ability of countries to manage and respond to disasters. | Highlights countries with weak emergency response systems, governance issues, and inadequate resources for immediate disaster response.<br><br>Guides the development of training programs, emergency preparedness plans, and investment in response infrastructure. |
| Lack of Adaptive Capabilities by Country | To show the long-term adaptive capacities of countries to future hazards. | Indicates which countries lack the ability to adapt to changing conditions and future risks.<br><br>Highlights the need for sustainable development, environmental management, and long-term planning.<br><br>Guides investments in education, innovation, and infrastructure to enhance adaptive capacities. |

**3. Scatter plots**

Scatter plots display the variation of two different items. They are represented by a set of points on a two-dimensional graph, with each point standing for a single data point and its associated values for both variables. The proposed study used scatter plots to identify correlations between WRI and its components, suggesting potential causative factors contributing to overall risk. Understanding which components have the strongest correlation with WRI can help prioritize interventions and resource allocation.

**4. Correlation heat map**

The correlation coefficients between several variables in a dataset can be visualized using a correlation heat map. It is a tool for analyzing correlation and regression patterns. The correlation coefficients are color-coded in the heat map, with each color signifying a different strength and direction of the correlation. The proposed study used these correlation plots to visually identify and quantify the strength and direction of relationships between the WRI score and other variables: exposure, vulnerability, susceptibility, lack of coping capabilities, and lack of adaptive capabilities. This helps to determine which factors have the strongest correlations with the WRI score, highlighting key areas that influence overall disaster risk.

**5. Heat map**

A heat map shows how numerous instances of a variable occur within each group. In a heat map, the colors denote either the values or the frequencies of the variable being mapped. The reason for using heat maps in the proposed study is to focus on the top ten countries with the highest WRI scores. This can highlight areas that are most at risk and require immediate attention.

**6. Line chart**

A line chart visualization connects data points with lines to show trends and changes over continuous intervals, typically time, making it ideal for tracking performance, identifying patterns, and forecasting.

By utilizing these diverse data visualization techniques, we were able to gain a comprehensive understanding of how the World Risk Index has changed over time for countries at both extremes of the risk spectrum. This analysis not only highlights temporal trends but also provides insights into the stability and variability of risk levels across different countries and periods.

## 3. RESULTS AND DISCUSSION

Firstly, we calculated the descriptive statistics of WRI, exposure, vulnerability, susceptibility, lack of coping capabilities, and lack of adaptive capacities and the statistics are depicted in Figure 3.

| | Year | WRI | Exposure | Vulnerability | Susceptibility | Lack of Coping Capabilties | Lack of Adaptive Capacities |
|---|---|---|---|---|---|---|---|
| **count** | 5018.000000 | 5018.000000 | 5018.000000 | 5018.000000 | 5018.000000 | 5018.000000 | 5018.000000 |
| **mean** | 2012.500000 | 8.202599 | 5.326441 | 24.106495 | 22.527672 | 18.743826 | 42.899163 |
| **std** | 7.500747 | 9.228445 | 10.049251 | 16.495837 | 16.769253 | 20.326771 | 18.066753 |
| **min** | 2000.000000 | 0.160000 | 0.020000 | 1.330000 | 1.470000 | 0.090000 | 2.270000 |
| **25%** | 2006.000000 | 2.230000 | 0.240000 | 11.880000 | 9.772500 | 6.142500 | 30.570000 |
| **50%** | 2012.500000 | 4.190000 | 1.050000 | 19.620000 | 16.735000 | 11.305000 | 44.405000 |
| **75%** | 2019.000000 | 10.680000 | 5.237500 | 30.547500 | 31.080000 | 14.790000 | 56.957500 |
| **max** | 2025.000000 | 47.420000 | 64.590000 | 75.730000 | 76.910000 | 81.490000 | 84.840000 |

**Figure 3: Descriptive statistics of WRI, exposure, vulnerability, susceptibility, lack of coping capabilities, and lack of adaptive capacities.**

**Table 2: Statistics by WRI vs. country**

| | Maximum scores | | Minimum scores | |
|---|---|---|---|---|
| **Factor** | **Score** | **Country** | **Score** | **Country** |
| WRI | 47.42 | Philippines | 0.16 | Monaco |
| Exposure | 64.59 | Andorra | 0.02 | China |
| Vulnerability | 75.73 | Somalia | 1.33 | Monaco |
| Susceptibility | 76.91 | Central African Republic | 1.47 | Monaco |
| Lack of Coping Capabilities | 81.49 | Somalia | 0.09 | Finland |
| Lack of Adaptive Capabilities | 84.84 | South Sudan | 2.27 | Bahrain |

According to Figure 3, the maximum WRI score is 47.42, and the minimum WRI score is 0.16. The extensive range of WRI scores, from 0.16 to 47.42, highlights the considerable variability in risk levels between countries in the dataset. Some nations mentioned in the dataset are almost completely protected from natural disasters or have exceptional risk management systems, with WRI scores closer to the low score of 0.16, while others are at high risk, with WRI scores closer to the high score of 47.42.

In addition, Table 2 lists the countries that recorded the highest and lowest scores for each WRI-based metric, considering all data recorded over the years. Monaco has a WRI score of 0.16, meaning the region is almost completely protected from natural disasters or has exceptional risk management systems as of now. The WRI score of the Philippines is 47.42, indicating that this region is at high risk and not well protected from natural disasters.

Furthermore, a cat plot graph was generated to observe how the WRI, exposure, vulnerability, susceptibility, lack of coping capabilities, and lack of adaptive capabilities have varied over the recorded years, as depicted in Figure 4. WRI and exposure have varied identically over the years. Vulnerability and susceptibility have also differed similarly, having dropped in 2009, increasing thereafter. Lack of adaptive capabilities have been observed over the years with irregularities.

Moreover, Figure 5 depicts a line plot that shows the behavior of the top countries over the 25 years until 2025. The Philippines has been recorded with the highest WRI, indicating the risk factor for the country. Having started with the highest WRI score in 2000, however, Mexico has seen its scores degrade over time. India also has a downward trend starting from 2018, stabilizing from 2020 onwards. Guatemala and Vanuatu illustrate similar behavior. Overall, after 2020, the WRI scores show a stabilizing trend till 2025.

Furthermore, it is vital to observe the distribution of WRI, exposure, vulnerability, and susceptibility scores over the regions in the year 2025. To visualize this, Choropleth maps were used and shown in Figure 6. When considering Choropleth maps, it is clear that vulnerability and susceptibility are higher in the regions of Africa and Asia. All over the world, the WRI score shows moderate values.

As this study's main field of interest is the WRI score, it is vital to illustrate the relationship of the WRI score with other scores. For this purpose, scatterplots were generated, as in Figure 7.

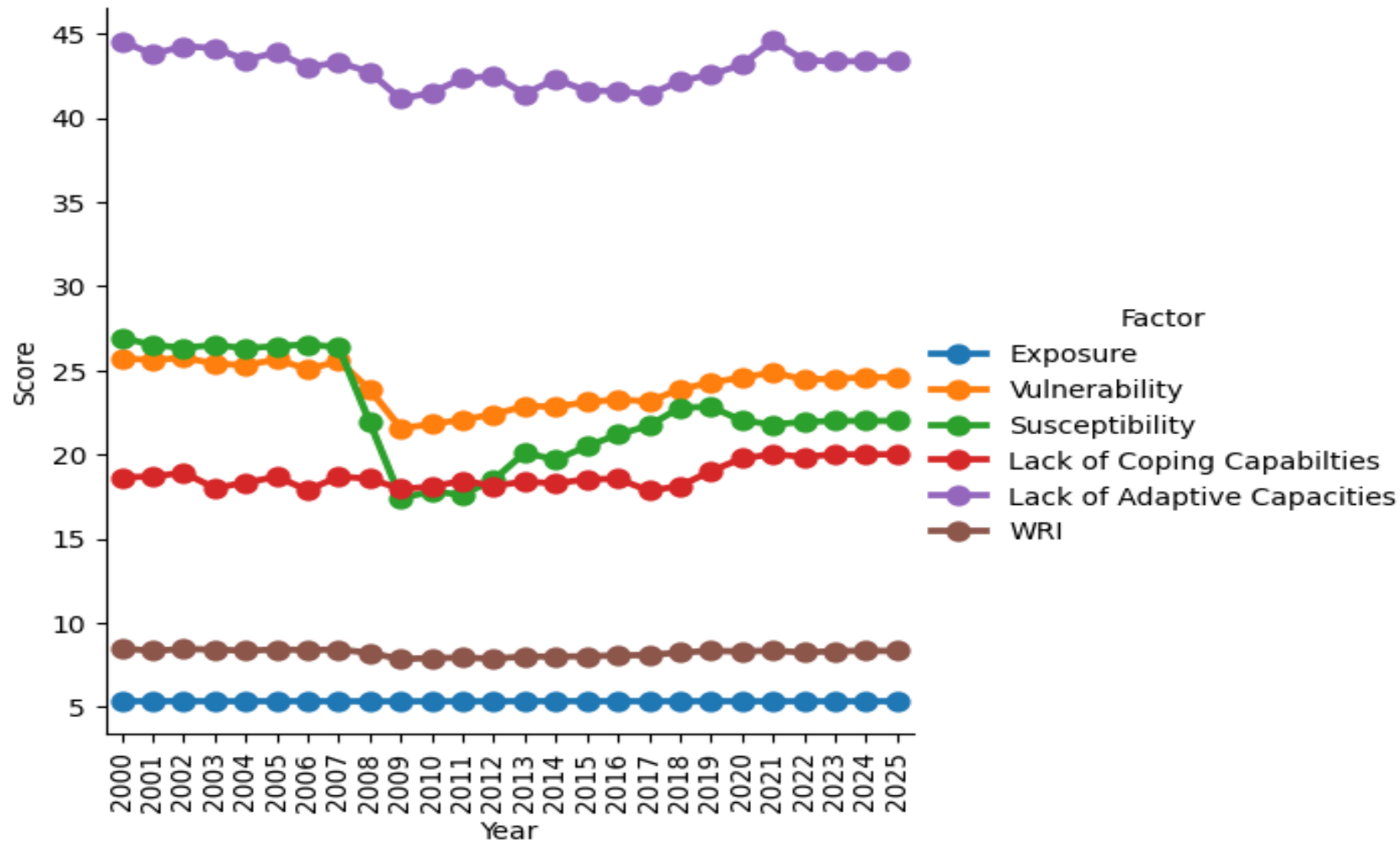

**Figure 4: Behavior of WRI, exposure, vulnerability, susceptibility, lack of coping capabilities, and lack of adaptive capabilities over the years.**

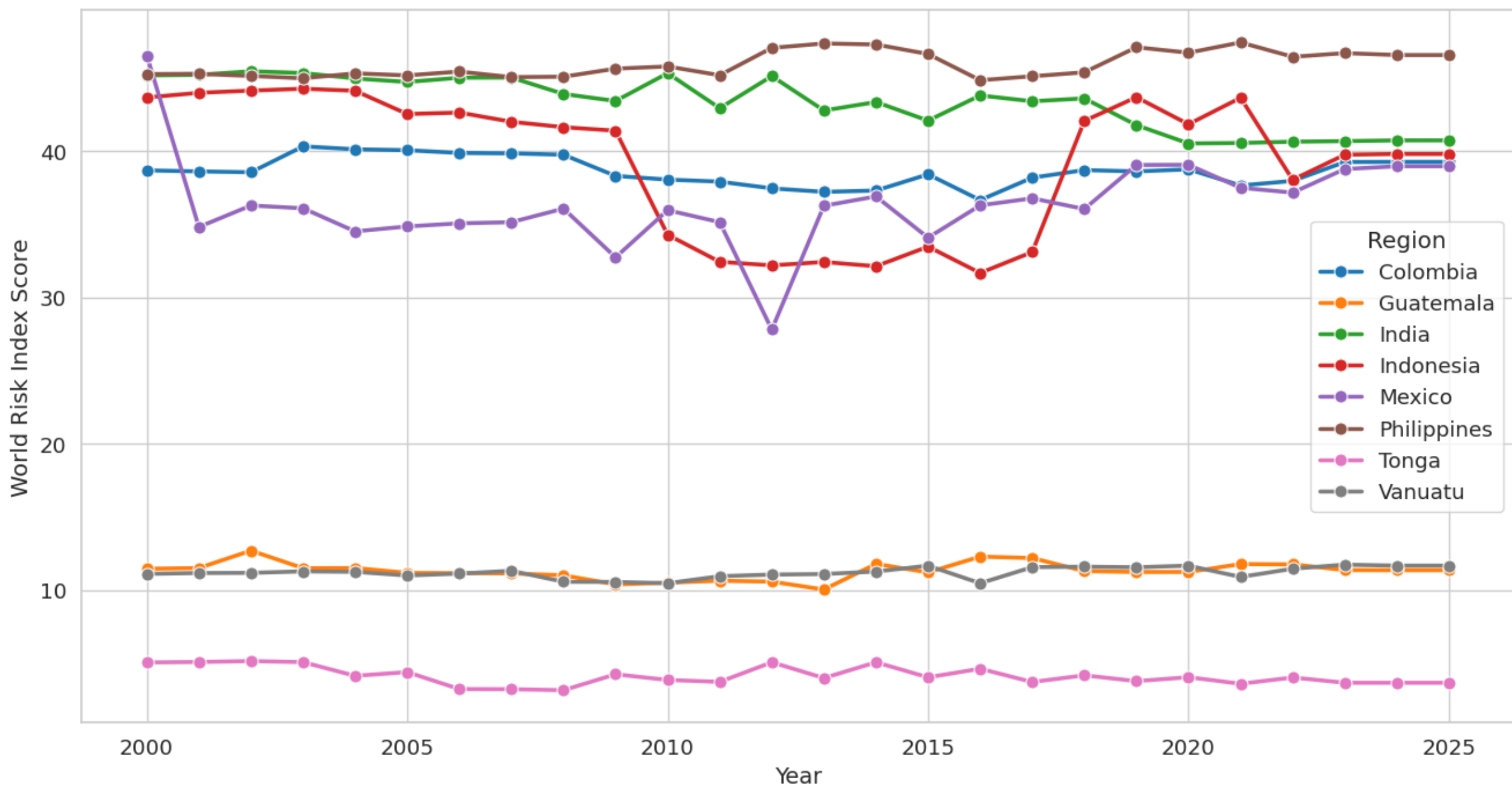

**Figure 5: Behavior of WRI of riskiest countries over the years.**

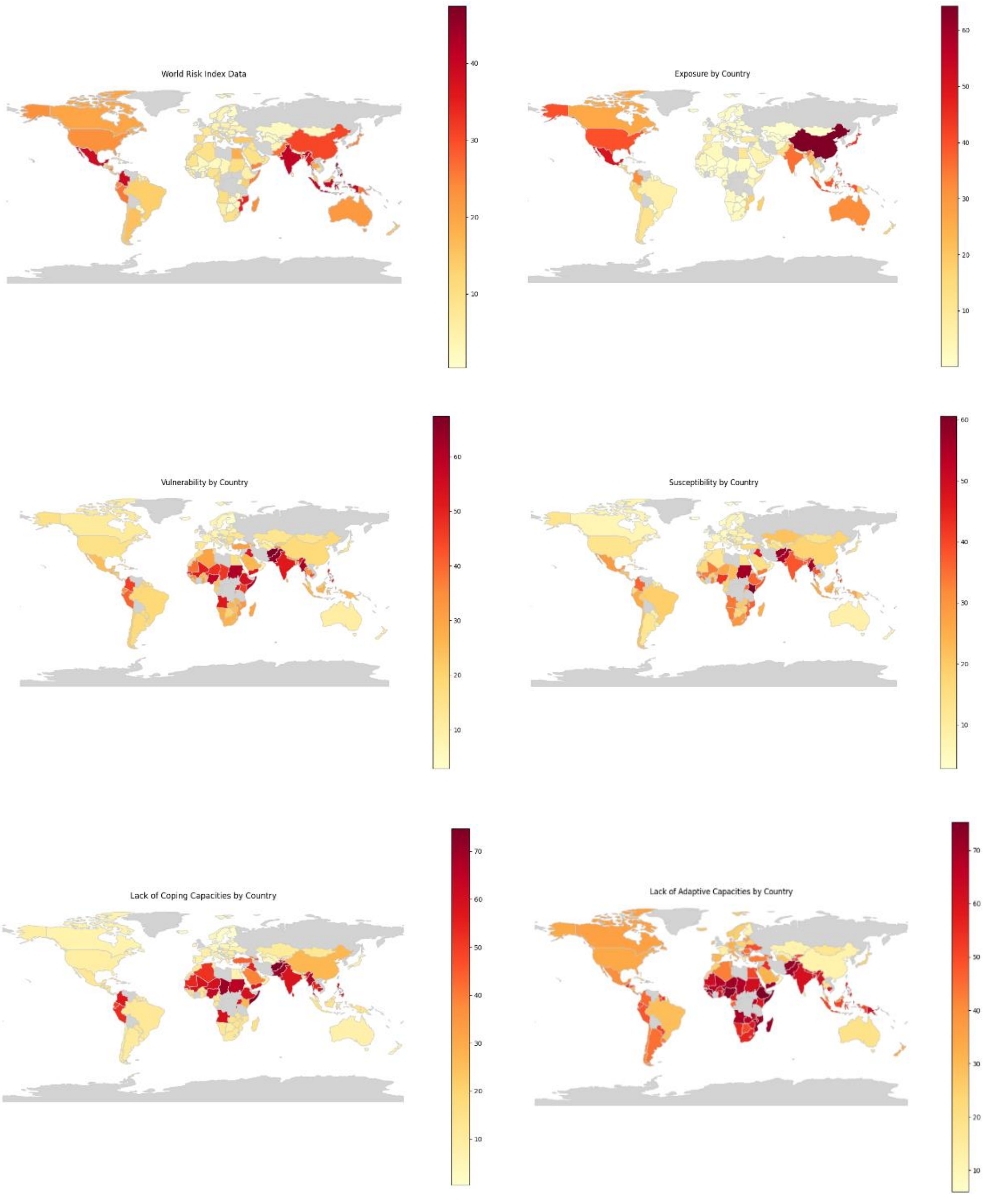

**Figure 6: Geographical illustration of WRI, exposure, vulnerability, and susceptibility of countries according to scores.**

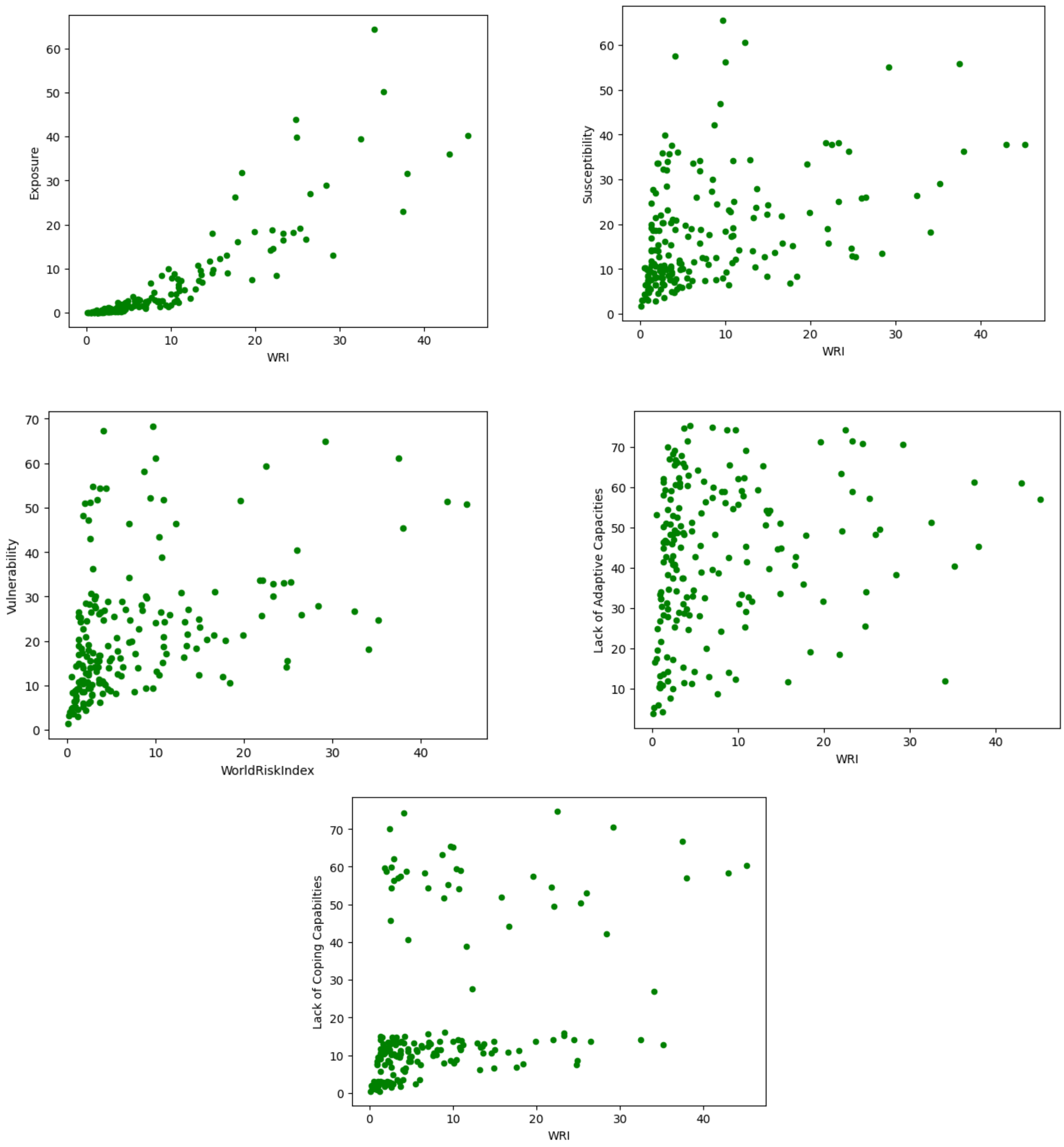

**Figure 7: Scatter plots illustrating the relationship between WRI and other scores.**

When considering the scatter plots depicted in Figure 7, exposure has a clear positive correlation with the WRI score. However, other factors of vulnerability, susceptibility, lack of coping capabilities, and lack of adaptive capabilities do not have a clear relationship with the WRI score.

However, to quantify these relationships, the correlation between the WRI and other variables was calculated and illustrated by a correlation plot as in Figure 8.

According to the correlation plot, it can be further noted that exposure has the highest correlation with the WRI score.

Furthermore, it is important to examine the behavior of the WRI score of each country over the years. Bar graphs in descending order, as shown in Figure 9, were used to visualize these scores.

According to Figure 9, it is evident that the Philippines has been recorded with the highest WRI in most years, indicating that the country is prone to risk, as deduced by previous plots as well.

Furthermore, a heat map as shown in Figure 10 was constructed to detect the occurrence of countries in the top ten with the highest WRI scores. It is notable that countries such as Egypt, Iran, Turkey, and Panama have only a few occurrences over the years, while other countries have been recorded more frequently over the time, showing that the risky behavior has been logged throughout the time.

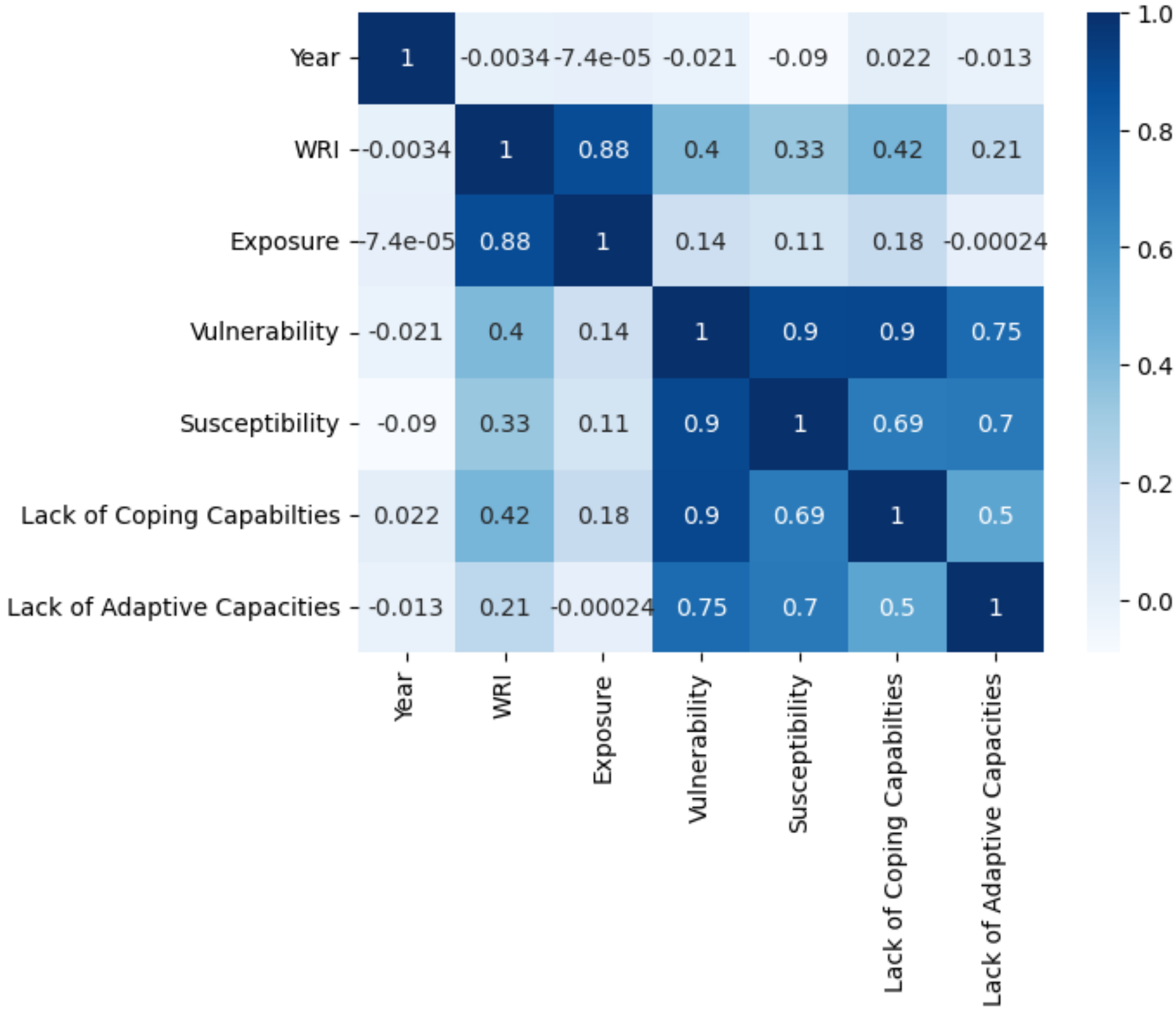

**Figure 8: Correlation plot describing the relationship between WRI score and other variables.**

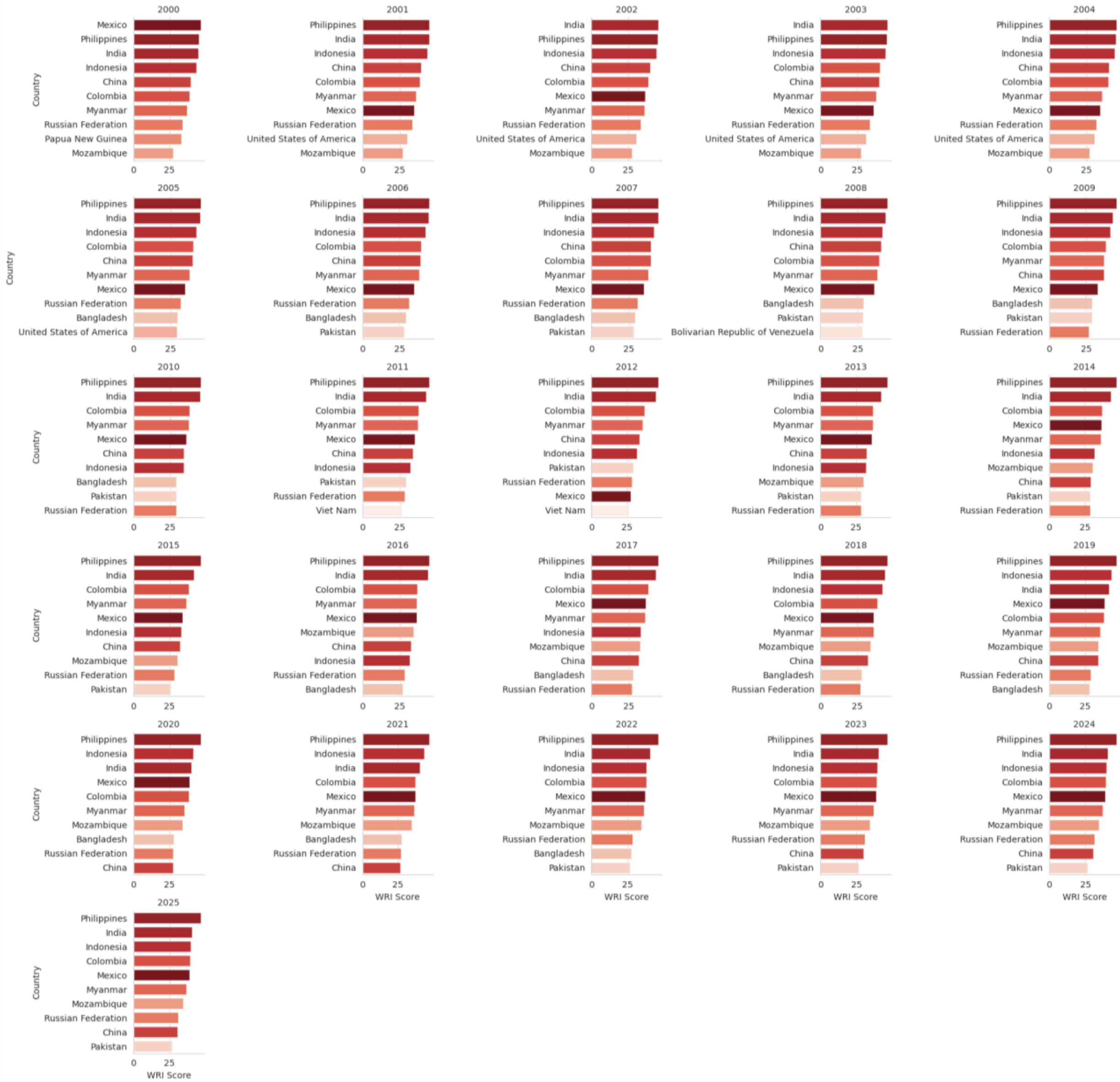

**Figure 9: WRI scores of countries over the years in descending order.**

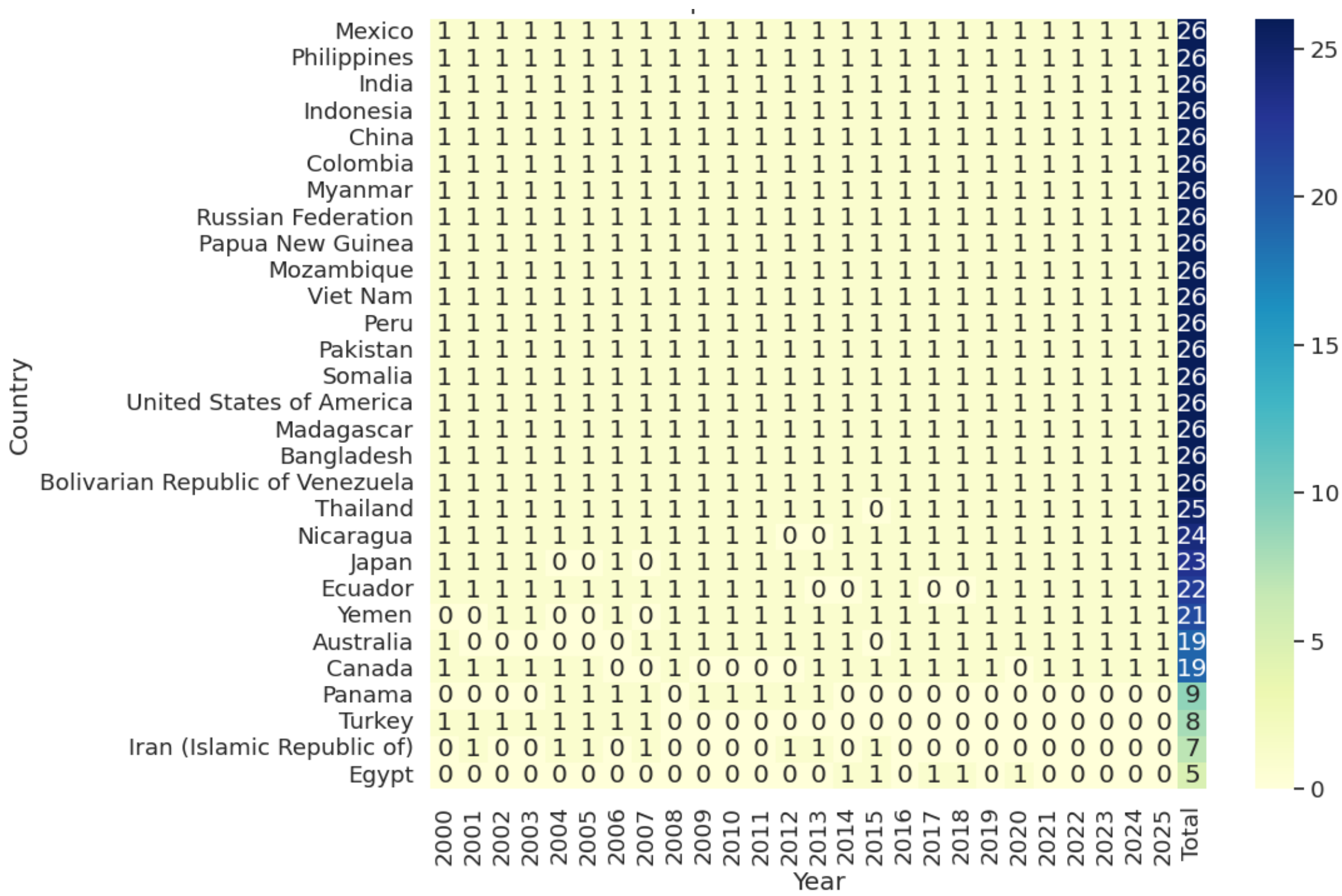

**Figure 10: Heat map illustrating the total number of counts of how many times a country has appeared in the top ten with the highest WRI scores across all years.**

Selecting the most suitable visualization technique depends on the specific goals, characteristics of the data, and the audience. Table 3 shows an ordered list of the most suitable visualization techniques for analyzing the World Risk Index using data visualization techniques.

Moreover, analyzing the World Risk Index dataset using data visualization techniques requires choosing the most efficient and usable visualization libraries. The comparison of some popular libraries used in the proposed study is mentioned in the Table 4 based on their efficiency and usability for analyzing WRI data.

The proposed study suggested that one needs to consider the type of plots required, the level of interactivity needed, and the familiarity with the syntax when choosing a visualization library for analyzing WRI data to ensure efficient and effective exploration and communication of data.

Table 5 shows the libraries used in the proposed study for time analysis, along with their application to efficiency, usability, and the types of plots used.

## 4. CONCLUSION

This study utilized the World Risk Index (WRI) data to analyze the risk levels calculated for 193 countries that had data from 2000 - 2025, discussing various WRI components and investigating diverse data analysis and visualization methods. Various visualizations, including catplots, Choropleth maps, scatter plots, correlation heat maps, bar graphs, and heat maps, were employed to identify important details. The comprehensive analysis and visualization techniques allowed us to understand trends over time, identify countries with the highest and lowest risk levels, and gain insights into the WRI scores. However, this study could be further enhanced by incorporating additional indices such as the Global Climate Index and Information Index. This would provide a more holistic understanding of global risk factors and enhance the effectiveness of disaster risk reduction strategies.

**Table 3. The most suitable visualization techniques for analyzing the World Risk Index using data visualization techniques**

| **Visualization technique** | **Factors that affect analyzing the World Risk Index (WRI) using data visualization techniques** |
|---|---|
| Choropleth maps | Offers a geospatial perspective of metrics, making it easy to identify high-risk areas and regional disparities.<br>Effective for communicating global risk distributions and highlighting hotspots. |
| Heat map | Provides a comprehensive overview of scores and the occurrence of high-risk countries over multiple years.<br>Useful for identifying trends, patterns, and relationships within the data. |
| Correlation heat map | Helps analyze relationships between WRI scores and other variables.<br>Provides insights into factors influencing disaster risk and informs policy-making. |
| Cat plot (Categorical Plot) | Useful for comparing categorical variables such as WRI scores across years.<br>Provides insights into how WRI scores vary across years. |
| Scatter plots | Useful for visualizing relationships between WRI scores and other continuous variables.<br>Helps identify correlations and trends between WRI scores and variables. |
| Line chart | Useful to observe the behavior of data over a wide range of years |
| Bar graphs | Allows for easy identification of countries with the highest WRI scores, prioritizing resource allocation and intervention efforts.<br>Highlights countries most in need of assistance and policy attention |

**Table 4. Popular libraries used in the proposed study based on their efficiency and usability for analyzing WRI data**

| **Library used in the proposed study** | **Efficiency** | **Usability** |
|---|---|---|
| Matplotlib | Highly efficient for creating basic plots and offers a wide range of customization options. | While powerful and has a steeper learning curve compared to other libraries. However, its syntax is similar to MATLAB, making it familiar to users with MATLAB experience. |
| Seaborn | Builds on top of Matplotlib and provides a high-level interface for creating attractive statistical visualizations with minimal code. | Seaborn's syntax is intuitive and user-friendly, making it easy to create complex plots like heat maps, violin plots, and categorical plots. |
| Geopandas | Efficient for working with geospatial data and creating Choropleth maps directly from shapefiles or GeoDataFrames. | Provides a familiar interface for working with spatial data in a tabular format, making it accessible to users familiar with Pandas. |

**Table 5: Libraries used for the proposed study for time analysis**

| **Library** | **Efficiency** | **Usability** | **Used for time analysis** |
|---|---|---|---|
| Pandas | High-performance data manipulation | Easy to use with simple syntax | Data loading, manipulation, and preparation |
| Matplotlib | Efficient plotting with customizable options | Versatile and widely used plotting library | Line plot, Histogram plot, Box and whisker plot |
| Seaborn | Built on Matplotlib, with enhanced efficiency | High-level interface for attractive plots | Box and whisker plot, Density plot |

## 6. REFERENCES

ArcGIS (n.d.) 'Create and use a choropleth map', *ArcGIS Insights Documentation*. https://doc.arcgis.com/en/insights/latest/create/Choropeth-maps.htm

Fayaz, M. *et al.* (2018) 'A blended risk index modeling and visualization based on hierarchical fuzzy logic for water supply pipelines assessment and management,' *Processes*, 6(5), p. 61. https://doi.org/10.3390/pr6050061.

Humanitarian Data Exchange (2025) WorldRiskIndex. https://data.humdata.org/dataset/worldriskindex.

Peduzzi, P. *et al.* (2009) 'Assessing global exposure and vulnerability towards natural hazards: the Disaster Risk Index,' *Natural Hazards and Earth System Sciences*, 9(4), pp. 1149–1159. https://doi.org/10.5194/nhess-9-1149-2009.

Poljansek, K., Marín Ferrer, M., De Groeve, T. and Clark, I. (2017) Science for disaster risk management 2017: knowing better and losing less. Zurich: ETH Zurich.

Population Matters (2024) *Population: the numbers - Population Matters.* https://populationmatters.org/the-facts-numbers.

Sadiku, M., Shadare, A.E., Musa, S.M., Akujuobi, C.M. and Perry, R. (2016) 'Data visualization',

*International Journal of Engineering Research and Advanced Technology (IJERAT)*, 2(12), pp. 11–16.

Suhaila, J. (2021) 'Functional data visualization and outlier detection on the anomaly of El Niño Southern oscillation,' *Climate*, 9(7), p. 118. https://doi.org/10.3390/cli9070118.

Welle, T. and Birkmann, J. (2015) 'The World Risk Index – an approach to assess risk and vulnerability on a global scale,' *Journal of Extreme Events*, 02(01), p. 1550003.
https://doi.org/10.1142/s2345737615500037.